\pdfoutput=1

\documentclass[11pt]{article}

\usepackage{acl}

\usepackage{times}
\usepackage{latexsym}

\usepackage[T1]{fontenc}

\usepackage[utf8]{inputenc}

\usepackage{microtype}

\usepackage{inconsolata}

\usepackage{microtype}

\usepackage{graphicx}
\usepackage{enumitem}
\usepackage{supertabular,booktabs}
\usepackage{tabularx}
\usepackage{xcolor}
\usepackage{color, colortbl}
\usepackage[explicit]{titlesec}
\usepackage{makecell}
\usepackage{multirow}
\usepackage{pifont}
\usepackage[caption=false]{subfig}
\usepackage[noorphans,vskip=0ex]{quoting}
\usepackage{amsmath}
\usepackage{amssymb}
\usepackage{listings}
\usepackage{mdframed}
\usepackage{tikz}
\usepackage[most]{tcolorbox}
\usepackage{footnote}

\titlespacing{\paragraph}{%
  0em}{
  0\baselineskip}{
 0\baselineskip}%
\usepackage{pdfpages}
\usepackage{float}
\usepackage[caption = false]{subfig}

\renewcommand{\vec}[1]{\mathbf{#1}}

\renewcommand{\paragraph}[1]{\vspace{0.2cm}\noindent\textbf{#1}}

\title{Leveraging Contextual Information for Effective Entity Salience Detection}

\author{
Rajarshi Bhowmik \qquad Marco Ponza \qquad Atharva Tendle \qquad Anant Gupta \\
\stepcounter{footnote}\textbf{Rebecca Jiang}~\Thanks{Work was done while the author was affiliated with Bloomberg} \qquad \textbf{Xingyu Lu} \qquad \textbf{Qian Zhao} \qquad \textbf{Daniel Preo\c{t}iuc-Pietro}
\\
Bloomberg\\
\footnotesize
\{\href{mailto:rbhowmik6@bloomberg.net}{\texttt{rbhowmik6}}, \href{mailto:dpreotiucpie@bloomberg.net}{\texttt{dpreotiucpie}}\}{\texttt{@bloomberg.net}}
}

\begin{document}
\maketitle

\begin{abstract}

In text documents such as news articles, the content and key events usually revolve around a subset of all the entities mentioned in a document. These entities, often deemed as salient entities, provide useful cues of the aboutness of a document to a reader. Identifying the salience of entities was found helpful in several downstream applications such as search, ranking, and entity-centric summarization, among others. Prior work on salient entity detection mainly focused on machine learning models that require heavy feature engineering. We show that fine-tuning medium-sized language models with a cross-encoder style architecture yields substantial performance gains over feature engineering approaches. To this end, we conduct a comprehensive benchmarking of four publicly available datasets using models representative of the medium-sized pre-trained language model family. Additionally, we show that zero-shot prompting of instruction-tuned language models yields inferior results, indicating the task's uniqueness and complexity.

\end{abstract}

\section{Introduction}

Many NLP studies have highlighted the importance of entities to understanding the semantics of a document~\cite{wu-et-al-2020, meij2012}. Automatically identifying entities in unstructured text documents and linking them to an underlying knowledge base, such as Wikipedia, is one of the core NLP tasks, with multiple shared tasks~\cite{conll-2003,strauss2016results}, benchmarks~\cite{hoffart-etal-2011-aida-conll,hovy-etal-2006-ontonotes,pradhan-etal-2013-towards,rijhwani-preotiuc-pietro-2020-temporally,derczynski-etal-2016-broad}, and studies~\cite{kolitsas-etal-2018-e2e-el, nguyen-2014-aida-light} dedicated to solving them.

\begin{figure}[t!]
    \centering
 \includegraphics[width=.80\linewidth]{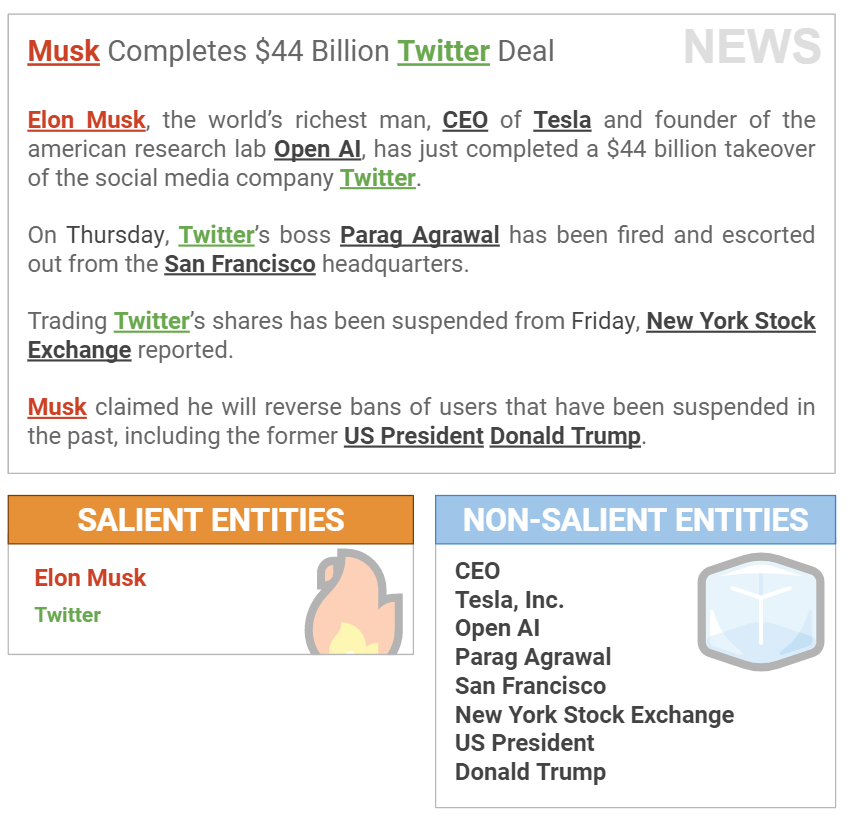}
\caption{An example of a document with salient and non-salient entities. Entity mentions are highlighted in text.}
\label{f:examples}
\end{figure}

Although an entity may play a crucial semantic role in document understanding, not all entities in a text document play equal roles. Some entities are the central subjects or actors within a document, around which the content and the key events revolve. Others are mentioned only to provide additional context to the main event. For example, some entities may be actors in peripheral events, while others are deemed uninformative to the understanding of the document. Thus, \textit{entity salience} in a text is defined as a binary or ordinal rating to quantify the extent to which a target entity is central to a given piece of text~\cite{gamon2013,dunietz-gillick-2014-new}. Figure~\ref{f:examples} provides an example text along with the mentioned entities and their salience. We note that the salience of an entity to a text is independent of the user's interest when reading or searching the document~\cite{gamon2013}, which is usually referred to as \textit{entity relevance}. It is also distinct from \textit{entity importance}, which quantifies the overall importance of the entity independent of the document. Automatically inferring entity salience was shown to aid search~\cite{gamon2013}, improve ranking results~\cite{xiong2018towards}, entity detection~\cite{trani-2018-sel}, and enable entity-centric applications such as entity-centric summarization~\cite{maddela-etal-2022-entsum}.

In this paper, we study the effectiveness of Transformer-based Pre-trained Language Models (PLMs) in the task of entity salience detection. Prior work on determining entity salience relied on heavy feature engineering to craft features explicitly covering relevant aspects, such as entity frequency~\cite{dunietz-gillick-2014-new,dojchinovski-etal-2016-crowdsourced}, position of entity mentions within a document~\cite{dunietz-gillick-2014-new,trani-2018-sel}, relations to other entities~\cite{trani-2018-sel}, document features, such as its length~\cite{gamon2013} and lexical features, such as the name of the entity or its context. Only a single recent work attempted to use PLMs in a pipeline which included key entity detection, albeit the scope of the evaluation was limited to a single high performing dataset~\cite{zhao2021bert}. In contrast, our proposed method uses a cross-encoder architecture where a target entity's name or alias and its contextual mentions in a text document are encoded by a PLM encoder. The classifier uses the contextual representation and, optionally, positional information about the entity encoded through the decile position embedding vector of mentions to determine the salience score of a target entity.

We conduct experiments on four publicly available datasets, two of which were human annotated and two that were curated semi-automatically. We fine-tune several cross-encoders using PLMs and demonstrate that these yield consistent and significant improvements over feature-based methods, as well as prompting instruction-tuned PLMs. The latter shows the novelty and complexity of the task of entity salience detection, which requires the model to learn significant task-specific semantic knowledge for this natural language understanding task.

Our contributions in this paper are the following:
\begin{itemize}[noitemsep,topsep=0pt,leftmargin=1em]
    \item We propose a cross-encoder style architecture with explicit encoding of position information for entity salience detection that shows consistent improvements of 7 -- 24.4 F1 scores over previous feature engineering approaches.
    \item We establish a uniform benchmark of two human annotated and two semi-automatically curated datasets for the task of entity salience detection that we expect to be beneficial to future study of this task;
    \item A faceted analysis of the models' predictive behaviour.
\end{itemize}

\section{Related Work}

Understanding the aboutness of a document is one of the long-standing goals of research in both Information Retrieval and Natural Language Processing~\cite{gamon2013}. Several types of approaches have been proposed, including extracting key-terms~\cite{hulth2003improved,mihalcea-tarau-2004-textrank}, identifying latent topics~\cite{blei2003latent}, or generating text summaries~\cite{erkan2004lexrank}. There has been a recent focus in using entities to understand the content of a document. Towards this goal, the task of entity salience has been first described for web pages in~\cite{gamon2013} and for news content in~\cite{dunietz-gillick-2014-new}. This task can be viewed as a restricted form of keyword or keyphrase extraction~\cite{alami2020automatic} if salience is binary. For the rest of this study, we will use the concept of salience as described in~\cite{gamon2013}.

The salience labels for entities were obtained either by crowdsourcing labels from multiple raters to identify salient entities~\cite{gamon2013, dojchinovski-etal-2016-crowdsourced, trani-2018-sel, maddela-etal-2022-entsum} or by using proxies. For example, \cite{dunietz-gillick-2014-new} hypothesize that salient entities are those that appear in the article's abstract. \cite{wu2020wn} identifies an entity as salient if the Wikinews category
that corresponds to the entity is also labeled as the category of the article.

Past studies mostly proposed machine learning methods to infer the salience of a given entity that relied on hand-crafted features. Features that can be computed from the target entity mentions and document alone can be categorized into the following: positional (e.g., position in the document, if entity is in the abstract)~\cite{dunietz-gillick-2014-new}, count-based (e.g., number of references to the entity)~\cite{dunietz-gillick-2014-new, wu2020wn}, local context~\cite{trani-2018-sel}, or global context~\cite{ponza-2018-swat}. Further, joint entity salience resolution can be performed by creating features using the entity graph (e.g., centrality in the entity graph)~\cite{dunietz-gillick-2014-new,trani-2018-sel}. Finally, past work also showed that incorporating external knowledge about entities from knowledge bases can boost predictive performance~\cite{dojchinovski-etal-2016-crowdsourced}.

Automatically inferring salience for entities can directly benefit multiple downstream applications, such as improving ranking results for queries containing entities~\cite{xiong2018towards} or improving the performance of entity detection by joint modelling~\cite{trani-2018-sel}.
Moreover, by inferring salience, new entity-centric applications can be built, such as highlighting salient entities in search~\cite{gamon2013}, improving the interpretability of news trends through salient entities~\cite{ponza2021contextualizing}, or identifying entities for creating entity-centric summaries of news stories~\cite{maddela-etal-2022-entsum,hofmann-coyle-etal-2022-extractive}.

\section{Problem Definition}

We use the concept of salience as introduced in~\cite{gamon2013}: salient entities are entities explicitly mentioned in the document that are objectively important as a function of the structure of the text. 

The goal of the salience model is to produce a single salience score $\psi(e)$  for the entity $e$ using only the document $D$ and the explicit entity mentions $\mathcal{M}_e$. We consider using external knowledge, such as information about entities from knowledge bases, to be outside the scope and leave integration of such knowledge for future work.

\section{Methods}
 Pre-trained Language Models (PLMs) have shown a remarkable ability to encode syntactic and semantic knowledge in their parameters \cite{tenney2018you,tenney-etal-2019-bert} that can be leveraged when fine-tuned on downstream natural language understanding (NLU) tasks. We postulate that PLMs can be harnessed to help in entity salience detection, a target-based document-level NLU task. In this section, we present an architecture based on the  cross-encoder setup adapted to the task of entity salience detection.
 
\subsection{Cross-encoder} 
\label{sec:cross-encoder}
\begin{figure}[t!]
    \centering
\includegraphics[width=1.0\linewidth]{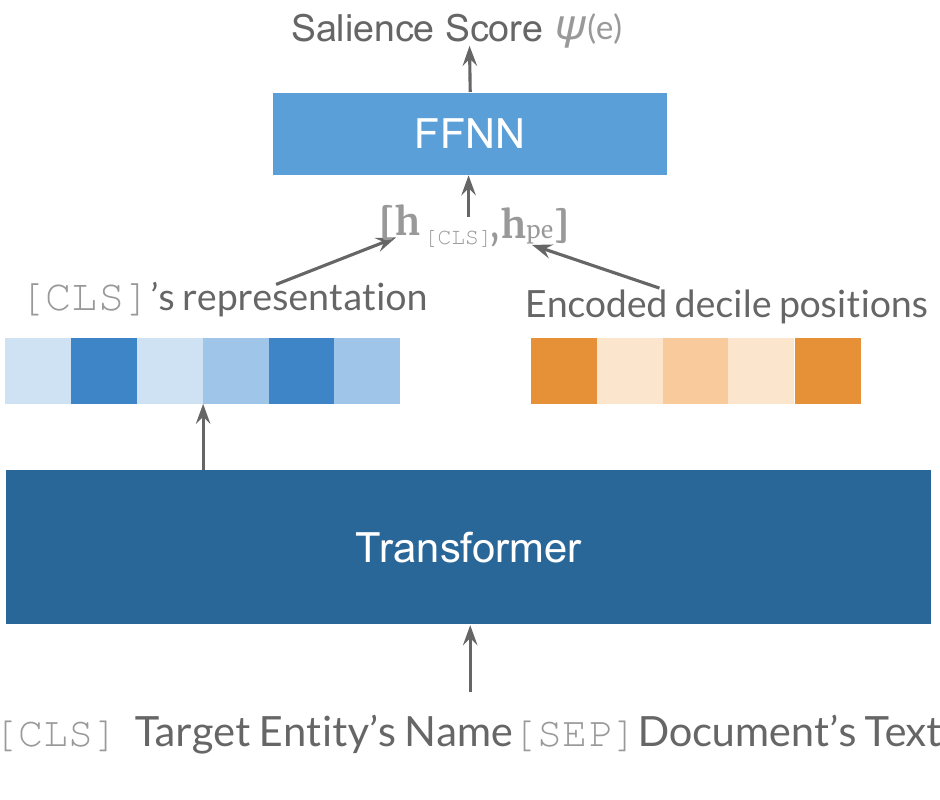}
\caption{Graphical representation  of the cross-encoder architecture with decile position encoding.}
\label{fig:cross-encoder}
\end{figure}

\paragraph{Encoding}
Given a document $D$ and a target entity $e$, which is mentioned in the document, we concatenate the target entity's name and the document using a special \texttt{[SEP]} token. We then encode the text using a Transformer-based pre-trained encoder. Figure \ref{fig:cross-encoder} shows the graphical representation of the cross-encoder model. This setup allows the model to have deep cross attention between the target entity and the entire document. Note that we use special marker tokens \texttt{[BEGIN\_ENTITY]} and \texttt{[END\_ENTITY]} around each mentions $m \in \mathcal{M}_e$ of entity $e$ in document $D$.

\paragraph{Position Encoding}
We compute the decile positions for each entity mention ($m \in \mathcal{M}_e$) in the document $D$ by taking a positional index $p_m \in \{0, 1, \dots, 9\}$, indicating which part of document the mention belongs to if the document is partitioned into $10$ equal chunks. Depending on the number and positions of the mentions, the vector can contain multiple non-zero values in the $p$ vector. For example, if an entity $e$ has $1$ mention in the first decile, $2$ in the second decile, and $1$ mention in the fifth decile, then the input to the positional encoder would be $p_m = [1, 1, 0, 0, 1, 0, 0, 0, 0, 0]$. Note that we do not capture the number of mentions in each decile in $p_m$. To obtain positional embeddings, we use an embedding layer that maps positional indices to a dense vector of dimension $d_{model}$, formally $\vec{h}_{pe}(m) = \texttt{Embedding}(p_m)$.

\paragraph{Scoring} The output representation of the \texttt{[CLS]} token is concatenated with the mean position embedding vector $\vec{h}_{pe}$ and fed to a scorer module that produces a salience score $\psi(e) \in [0, 1]$ for entity $e$. The salience scorer is a feed-forward network with a sigmoid scoring function head. Formally,

\begin{equation*}
    \psi(e) = \sigma(\texttt{FFN}(\vec{h}_{\texttt{[CLS]}} || \vec{h}_{pe}))
\end{equation*}

\subsection{Optimization} We fine-tune the model described above by minimizing the binary cross entropy loss that is calculated using the ground truth binary salience labels and the predicted salience score $\psi(e)$.

\section{Datasets}
\label{sec:datasets}
\begin{table*}[t!]
    \centering
    \resizebox{1.9\columnwidth}{!}{
    \begin{tabular}{l|c|c|c|c}
    \toprule
     \textbf{Dataset} & \textbf{NYT-Salience} & \textbf{WN-Salience} & \textbf{SEL} & \textbf{EntSUM} \\ \hline
     \# Docs & 110,463 & 6,956 & 365 & 693 \\
     Doc Length (avg chars) & 5,079 & 2,106 & 1,660 & 4,995 \\
     \# Unique entities & 179,341 & 23,205 & 6,779 & 7,854 \\
     \# Mentions & 4,405,066 & 145,081 & 19,729 & 20,784 \\
     \% Salient entities & 14\% & 27\% & 10\% & 39\% \\
     Ground-truth & Abstract Alignment & Category Alignment  & Human & Human \\
    \bottomrule
    \end{tabular}
    }
    \caption{Summary statistics and label collection methods for the datasets used in our experiments.}
    \label{tab:dataset_stats}
\end{table*}

In this section, we describe our entity salience benchmark, which consists of four datasets: two datasets were curated using semi-automated methods and two used human annotations. We provide summary statistics of these datasets and label collection methods in Table~\ref{tab:dataset_stats}.

\paragraph{NYT-Salience} This dataset is introduced in \cite{dunietz-gillick-2014-new} and is the largest dataset to date for entity salience detection. The dataset is curated with an assumption that salient entities are mentioned in the abstract of a news article in the NYT Corpus \cite{sandhaus2008new}. Entities and their mentions are identified using a classical NLP pipeline involving POS tagging, dependency parsing, and noun phrase extraction. Despite being large-scale, the automatic dataset creation process could introduce noise as corroborated by moderate agreement numbers with human annotators on a subset of the data. The dataset contains a binary salience label for each entity.

\paragraph{WN-Salience} Introduced in~\cite{wu2020wn}, this is another automatically curated dataset consisting of Wikinews articles. These are annotated with Wikinews categories by their authors. WN-Salience identifies salient entities by using the hypothesis that an entity is salient if the Wikinews category that corresponds to the entity is also labeled as a category of the article. Similar to NYT-Salience, this dataset has binary salience labels.

\paragraph{SEL} This is another dataset based on Wikinews released by~\cite{trani-2018-sel}. However, unlike WN-Salience, this dataset is human annotated, where multiple human annotators ranked the salience of entities into one of four categories. To conform with the binary labels of the other datasets, we map the 4 categories into binary labels of $\{0, 1\}$ by mapping the bottom two classes to not salient and the top two classes to salient.

\paragraph{EntSUM} This dataset was introduced in~\cite{maddela-etal-2022-entsum}. To construct this dataset, a randomly selected set of entities spanning a subset of $693$ articles from the NYT corpus were assigned salience labels by human annotators on a four-point scale, ranging between $[0, 3]$. For each document entity pair, two independent annotations were collected, which were increased up to $5$ in case of disagreements. If the average annotation score is greater than $1.5$ for an entity, it is assigned a positive salience label. 

\subsection{Data Enrichment with Inferred Mentions} \label{sec:inferred mentions}
Except for EntSUM, the datasets do not have explicit entity mention offsets as annotations, which are necessary for many feature-based approaches and to compute positional embeddings. While SEL contains only the mention surface texts per entity, NYT-Salience and WN-Salience only provide the start and end character indices (aka mention offsets) of the very first mention of an entity. To this end, we infer additional mentions of an entity within the text using a combination of Flair NER~\cite{akbik2019flair} and pattern matching. 

For SEL, since the mentions are available, we use a pattern matching approach to match the surface text of the mentions to infer mention offsets. For NYT-Salience and WN-Salience, we first use Flair NER to identify mentions of named entities in the text. We attempt to match these mentions to the first mention of each entity in the document provided in the respective datasets. Since the surface text of other mentions may differ from the first mention, we additionally use the overlap between a mention's surface text and the entity name as a candidate mention for that entity. Applying this approach, we infer additional mentions of an entity in the text and their offsets. While this process could introduce some noise, the overall quality of the datasets are enhanced through this process.

\section{Experiments}\label{sec:main-experiments}
We experiment on our entity salience benchmark with our proposed PLM-based method, other ML and heuristic-based approaches used in past research, as well as an instruction-tuned PLM.

\subsection{Data Splits} \label{sec:dataset splits}
Prior works \cite{dunietz-gillick-2014-new,trani-2018-sel,wu2020wn} use inconsistent (or not reported) train/validation/test splits. NYT-Salience and WN-Salience datasets are provided with train/test splits (but no validation), whereas SEL dataset is provided without any splits. This makes it hard to benchmark previous works with a fair comparison across models. To overcome this issue, we do a temporal split of NYT-Salience's and WN-Salience's original training sets into a new train/validation sets based on the publication time of the news stories, which provides a more realistic testing setup~\cite{huang-paul-2018-examining,rijhwani-preotiuc-pietro-2020-temporally}. We also perform a temporal split of SEL and EntSUM datasets into train/validation/test sets. Further details about the dataset splits are provided in Appendix~\ref{sec:dataset split details}. 

\renewcommand*{\arraystretch}{1.1}
\begin{table*}[t!]
    \centering
    \resizebox{2\columnwidth}{!}{
    \begin{tabular}{l|l|l|c|c|c|c|c|c}
    \toprule
    \multirow{2}{*}{\textbf{Source}} & \multirow{2}{*}{\textbf{Type}} & \multirow{2}{*}{\textbf{Method}} &  \multicolumn{3}{c|}{NYT-Salience} & \multicolumn{3}{c}{WN-Salience} \\
    \cline{4-9}
    & & & \textbf{P} & \textbf{R} & \textbf{F1} & \textbf{P} & \textbf{R} & \textbf{F1} \\
    \hline
    \cite{dunietz-gillick-2014-new} & Heuristic & First Sentence & 59.5	& 37.8 & 46.2 & -- & -- & -- \\
    \cite{dunietz-gillick-2014-new} & ML & Position \& Frequency & 59.3 & 61.3	& 60.3 & -- & -- & -- \\
    \cite{dunietz-gillick-2014-new} & ML & All Features & 60.5 & 63.5 & 62.0 & -- & -- & -- \\
    \cite{ponza-2018-swat} & ML & SWAT & 62.4 & 66.0 & 64.1 & -- & -- & -- \\
    \cite{wu2020wn} & Heuristic & First Sentence & 56.0 & 41.0	& 47.3 & 47.9 & 53.2 & 50.4 \\
    \cite{wu2020wn} & ML & Positional Feature & 19.0 & 41.3 & 26.0 & 29.1 & \textbf{78.9} & 42.5 \\
    \cite{wu2020wn} & ML & Features \& GBDT & 39.2 & 59.7 & 47.3 & 29.2 & 48.1 & 36.3 \\
    \hline
    \multirow{6}{*}{Our Implementations} & Heuristic & Positional Headline & 57.5 & 42.0 & 48.5 & 46.1 & 51.5	& 48.7 \\
     & Heuristic & Positional Headline \& Lead & 49.8 & 55.4 & 52.5 & 41.0 & 60.0 & 48.7 \\
     & Heuristic & Entity Frequency & 53.7 & 53.3 & 53.6 & 37.3 & 61.9 & 46.6 \\
     & ML & Features \& GBDT & 61.0 & 57.4 & 59.2 & 46.2 & 53.3 & 49.5 \\
     & PLM (RoBERTa) & Target Entity Masking & 64.6 & 50.2 & 56.5 & 57.0	& 65.4 & 60.9 \\
    \hline
    \multirow{4}{*}{Our Models} &  PLM (RoBERTa)& cross-encoder & 75.9 & 87.1 & 81.1 & 71.8 & 73.6 & 72.7 \\   
     &  PLM (DeBERTa)& cross-encoder & 77.5 & 87.4 & \textbf{82.1} & 71.5 & 78.3 & \textbf{74.8} \\
    &  PLM (RoBERTa)& cross-encoder w/ position emb. & \textbf{78.7} & 84.2 & 81.4 & 71.2 & 76.7 & 73.8 \\   
    &  PLM (DeBERTa)& cross-encoder w/ position emb. & 75.9 & \textbf{88.4} & 81.7 & \textbf{73.3} & 76.1 & 74.7 \\
    \bottomrule
    \end{tabular}
    }
    \caption{Results on the NYT-Salience and WN-Salience datasets. The ground-truth of these datasets was generated via abstract/category alignment. The top section presents results as originally reported in the source papers.}
    \label{tab:results-nyt-wn}
\end{table*}

\renewcommand*{\arraystretch}{1.1}
\begin{table*}[t!]
    \centering
    \resizebox{2\columnwidth}{!}{
    \begin{tabular}{l|l|l|c|c|c|c|c|c}
    \toprule
    \multirow{2}{*}{\textbf{Source}} & \multirow{2}{*}{\textbf{Type}} & \multirow{2}{*}{\textbf{Method}} &  \multicolumn{3}{c|}{SEL} & \multicolumn{3}{c}{EntSUM} \\
    \cline{4-9}
     &  &  & \textbf{P} & \textbf{R} & \textbf{F1} & \textbf{P} & \textbf{R} & \textbf{F1} \\
    \hline
    \cite{trani-2018-sel} & ML & SEL (w/ 5-fold cross val.) & 50.0 & 61.0 & 52.0 & -- & -- & -- \\
    \cite{ponza-2018-swat} & ML & SWAT (w/ 5-fold cross val.) & 58.0 & 64.9	& 61.2 & -- & -- & -- \\
    \hline
    \multirow{7}{*}{Our Implementations} & Heuristic & Positional Headline & 26.6 & 78.4 & 39.7 & 60.7 & 18.5 & 28.4 \\
     & Heuristic & Positional Headline \& Lead & 22.1	& \textbf{87.1} & 35.3 & 51.2 & 31.6 & 39.1 \\
     & Heuristic & Entity Frequency & 13.5 & 57.8 & 21.9 & 48.4	& 54.0 & 51.0 \\
     & ML & Features \& GBDT & 26.6 & 78.4 & 39.7 & 60.7 & 52.0 & 56.0 \\
     & ML & SEL GBDT & \textbf{71.1} & 47.8 & 57.1 & -- & -- & -- \\
     & PLM (RoBERTa) & Target Entity Masking & 36.3 & 13.8 & 20.0 & 63.0 & 41.7 & 50.2 \\
    \hline
    \multirow{4}{*}{Our Models} &  PLM (RoBERTa)& cross-encoder & 51.6 & 73.6 & 60.6 & 65.5 & 60.6 & \textbf{63.0} \\   
    &  PLM (DeBERTa)& cross-encoder & 64.1 & 73.6 & \textbf{68.5} & 64.9 & 59.2 & 61.9 \\
     &  PLM (RoBERTa)& cross-encoder w/ position emb. & 63.0 & 69.9 & 66.3 & 67.5 & 57.0 & 61.8 \\  
     &  PLM (DeBERTa)& cross-encoder w/ position emb. & 67.3 & 62.4 & 64.7 & \textbf{72.1} & 51.5 & 60.1 \\
    \bottomrule
    \end{tabular}
    }
    \caption{Results on the SEL and EntSUM datasets. The ground-truth of these datasets was generated via human annotation. The top section presents results as originally reported in the source papers.}
    \label{tab:results-sel-entsum}
\end{table*}

\subsection{Baselines}

First, we list all methods used in past research, for which we report the results from the original papers.

\begin{itemize}[noitemsep,topsep=0pt,leftmargin=1em]

    \item \textit{First Sentence}. Classifies an entity as salient if it appears in the first sentence of the document's body; used in both \cite{dunietz-gillick-2014-new} and \cite{wu2020wn}.

    \item \textit{Position \& Frequency~\cite{dunietz-gillick-2014-new}}. Feeds the first sentence index and the frequency features of an entity into a logistic regression model.

    \item \textit{All Features~\cite{dunietz-gillick-2014-new}}. Uses a series of features based on position, frequency, and PageRank signals fed into a logistic regression model.


    \item \textit{SEL~\cite{trani-2018-sel}}. Uses a combination of features based on position, frequency, and Wikipedia graph statistics fed into a Gradient Boosted Decision Tree algorithm implemented in sklearn \cite{scikit-learn}.

    
    \item \textit{SWAT~\cite{ponza-2018-swat}}. Uses a set of features similar to the SEL Method described above, with the addition of features based on entity embeddings. All features are fed into a Gradient Boosted Decision Tree algorithm implemented in XGBoost~\cite{chen2015xgboost}.

    \item \textit{Positional Feature~\cite{wu2020wn}}. Uses the index of the first sentence in which the entity is mentioned as a feature in a logistic regression model. This method provides best results on the WN Salience dataset in~\cite{wu2020wn}.

\end{itemize}

Next, we re-implement a set of common methods based on the above baselines in order to be able to test them on all four datasets. This ensures the evaluation is performed on the same experimental setup.

\begin{itemize}[noitemsep,topsep=0pt,leftmargin=1em]    

    \item \textit{Positional Headline}. Classifies an entity as salient whether it appears in the headline of the input document.
    
    \item \textit{Positional Headline \& Lead}. Classifies an entity as salient if it appears in the headline of the document or in the first sentence (lead sentence) of the document.
    
    \item \textit{Entity Frequency}. Classifies an entity as salient if they are more frequent than a given value. For each dataset, we calculated different thresholds and reported the best results. Thresholds can be found in the Appendix.

    \item \textit{Features \& GBDT}. This method uses the most common features from past works~\cite{dunietz-gillick-2014-new,wu2020wn,trani-2018-sel,ponza-2018-swat} --- i.e., entity's first sentence index, and entity frequency --- and feeds them into a GBDT model implemented using LightGBM~\cite{ke2017lightgbm}.

    \item \textit{SEL GBDT}. Follows the method from~\cite{trani-2018-sel} and uses sklearn's GBDT~\cite{scikit-learn} to train a model on the features provided with the SEL dataset.

    \item \textit{Target entity masking}. This method feeds the input to a Transformer-based encoder (RoBERTa-base) with the target entity mentions represented through a special mask token. The salience prediction is obtained by mean pooling the mask token representations and passing this through a feed-forward network.
    
    \item \textit{Zero-shot prompting}. We test instruction-tuned LLMs using zero-shot prompting. The prompt introduces the task description, followed by the input text and a target entity, and it asks a yes/no question. It expects the model to generate either 'Yes' or 'No' as an answer. The LLMs, already instruction-tuned on a large collection of NLU tasks, attempt to provide an answer based on the prompt, input text, and target entity. This family of models has been demonstrated to be robust and versatile on multiple benchmarks \cite{chung2022scaling}. We use \texttt{Flan-UL2 (20B)}~\cite{tay2023ul2} and \texttt{LLaMa 2-Chat (7B) }~\cite{touvron2023llama} for evaluation.

\end{itemize}

\subsection{Experimental Setup}

We use RoBERTa-base \cite{liu2019roberta} and DeBERTa-v3-base \cite{he2023debertav} as the base PLM for experiments. For each of these base models, we train both a cross-encoder model and a cross-encoder model augmented with decile positional embeddings. 

For training our proposed models, we use AdamW~\cite{loshchilov2018adamw} as the optimizer. We perform a hyperparameter search for learning rate using the following set of values: $\{0.001, 0.0005, 0.0002, 0.0001, 0.00005\}$. We train our models for a maximum of $10$ epochs with early stopping based on the validation set performance. We pick the best performing model checkpoints for each dataset based on the performance on the validation set. In Tables \ref{tab:results-nyt-wn} and \ref{tab:results-sel-entsum}, we report the performance of our models and the baselines using the standard classification metrics (i.e., Precision, Recall, and F1) on the positive (salient) class, following previous research on entity salience.

For training and inference of each Transformer-based model, we use a single NVIDIA V100 GPU with 32GB GPU memory, 4 CPUs, and 128 GB of main memory.

\subsection{Results}

In Tables \ref{tab:results-nyt-wn} and \ref{tab:results-sel-entsum}, we present the experimental results of the baselines and our proposed models on the four datasets described in Section \ref{sec:datasets}. 

\paragraph{Comparison with feature-based methods.}
We observe that the cross-encoder model significantly outperforms all baseline models in F1 score. It also yields better precision compared to the baselines for three of the four datasets. Only for the SEL dataset does the SEL GBDT model trained on publicly available pre-computed features produce a model with better precision than the cross-encoder. We observe that adding the decile positional embedding with cross-encoder improves the precision across all datasets, but also degrades the recall in every dataset except NYT-Salience.

The Target Entity Masking approach, which also leverages contextual information with a transformer-based model yields mixed results. Overall, the model is able to obtain better precision than the feature-based models for all datasets except SEL, but the model suffers from poor recall across all datasets, resulting in significantly worse F1 scores especially when compared to cross-encoder models. 

Our re-implementation of positional methods and GBDT methods are consistent with the performance reported in prior works. The variance in numbers can be attributed to the enrichment of datasets with inferred mentions (Section~\ref{sec:inferred mentions}) and the explicit train/dev/test data split used in our experiments (Section~\ref{sec:dataset splits}). 

\subsection{Zero-shot prompting of large language models} We formulate the problem of salience detection with zero-shot prompting as follows: given a definition of entity salience task and document text, we ask the model to generate a "yes" or a "no" if a particular entity is salient or not. We experimented with two open source models (\texttt{Flan-UL2 (20B)} and \texttt{LLaMa 2-Chat (7B)}) available on Hugging Face \footnote{\url{www.huggingface.com}}, and present the results in Table~\ref{tab:llm-experiment}.
To the best of our knowledge, this is the first evaluation of zero-shot prompting of instruction-tuned models for the entity salience detection task. We observe that the \texttt{LLaMa 2-Chat} model with 7 billion parameters fails to yield any meaningful results as it produces only positive labels for all data points (hence we observe 100\% recall). The \texttt{Flan-UL2} model is able to generate both positive and negative labels. However, the precision remains too low across datasets. We further discuss causes for this performance in the Appendix (Section~\ref{sec:zeroshot}), along with the implementation details. Overall, these experiments suggest that entity salience detection is a unique task that is not similar to any other tasks these two models are instruction tuned on. 

\begin{table*}[t!]
    \centering
        \resizebox{2\columnwidth}{!}{
            \begin{tabular}{l|c|c|c|c|c|c|c|c|c|c|c|c}
            \toprule
             \multirow{2}{*}{\textbf{Model}} & 
             \multicolumn{3}{c|}{NYT-Salience} &
             \multicolumn{3}{c|}{WN-Salience} &
             \multicolumn{3}{c|}{SEL} & 
             \multicolumn{3}{c}{EntSUM} \\
            \cline{2-13}
            & \textbf{P} & \textbf{R} & \textbf{F1} & \textbf{P} & \textbf{R} & \textbf{F1} & \textbf{P} & \textbf{R} & \textbf{F1} & \textbf{P} & \textbf{R} & \textbf{F1}\\
            \hline
            Cross-encoder (DeBERTa) & 77.5 & 87.4 & 82.1 & 71.5 & 78.3 & 74.8 & 64.1 & 73.6 & 68.5 & 64.9 & 59.2 & 61.9 \\
            Flan-UL2 & 31.1 & 72.4 & 43.5 & 30.7 & 90.1 & 45.9 & 16.7 & 98.3 & 28.5 & 27.6 & 83.6 & 41.5 \\
            LLaMa 2-Chat & 14.6 & 100.0 & 25.4 & 27.1 & 100.0 & 42.6 & 9.49 & 100.0 & 17.3 & 19.2 & 100.0 & 32.2 \\
            \bottomrule
            \end{tabular}
        }
    \caption{Performance comparison of cross-encoder model with zero-shot prompting of LLMs.}
    \label{tab:llm-experiment}
\end{table*}

\section{Analysis}\label{sec:analysis}

\renewcommand*{\arraystretch}{1.1}
\begin{table*}[t!]
    \centering
        \resizebox{2\columnwidth}{!}{
            \begin{tabular}{l|c|c|c|c|c|c|c|c|c|c|c|c}
            \toprule
             \multirow{2}{*}{\textbf{Model}} & 
             \multicolumn{3}{c|}{NYT-Salience} &
             \multicolumn{3}{c|}{WN-Salience} &
             \multicolumn{3}{c|}{SEL} & 
             \multicolumn{3}{c}{EntSUM} \\
            \cline{2-13}
            & \textbf{P} & \textbf{R} & \textbf{F1} & \textbf{P} & \textbf{R} & \textbf{F1} & \textbf{P} & \textbf{R} & \textbf{F1} & \textbf{P} & \textbf{R} & \textbf{F1}\\
            \hline
            Cross-encoder w/ first mention & 54.2 & 57.5 & 55.8 & 69.6 & 80.4 & 74.6 & 59.8 & 76.1 & 67.0 & 69.1 & 53.2 & 60.2 \\
            Cross-encoder w/ all mentions & 77.5 & 87.4 & 82.1 & 71.5 & 78.3 & 74.8 & 64.1 & 73.6 & 68.5 & 64.9 & 59.2 & 61.9 \\
            \bottomrule
            \end{tabular}
        }
    \caption{Performance comparison of cross-encoder models with only the first mention vs.\ all inferred mentions.}
    \label{tab:first_vs_all}
\end{table*}

In this section, we perform an analysis of model predictions in order to gain more insights into model behavior and understand potential avenues for further improvement. We thus break down performance by different factors including: the importance of inferring all entity mentions, the position of the first entity mention, and entity mention frequency.

\begin{figure*}[th!]
\subfloat[Performance with respect to the position of the mentions. There are no mentions outside of the context window for NYT.]{\includegraphics[scale=0.275]{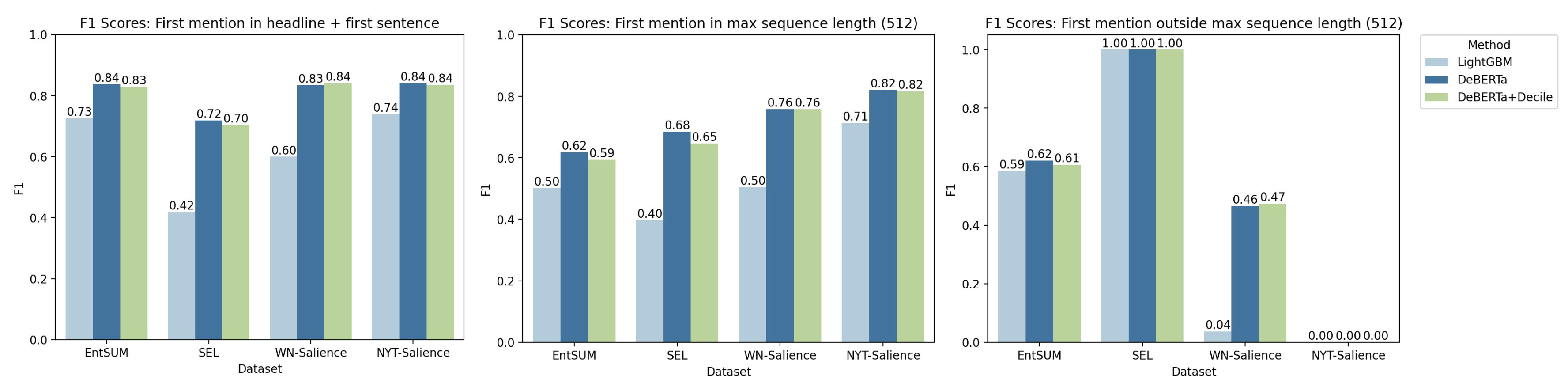} } \\
\subfloat[Performance with respect to the frequency of the entities. The test split of SEL dataset does not contain any entity with more than 10 mentions in a document.]{\includegraphics[scale=0.45]{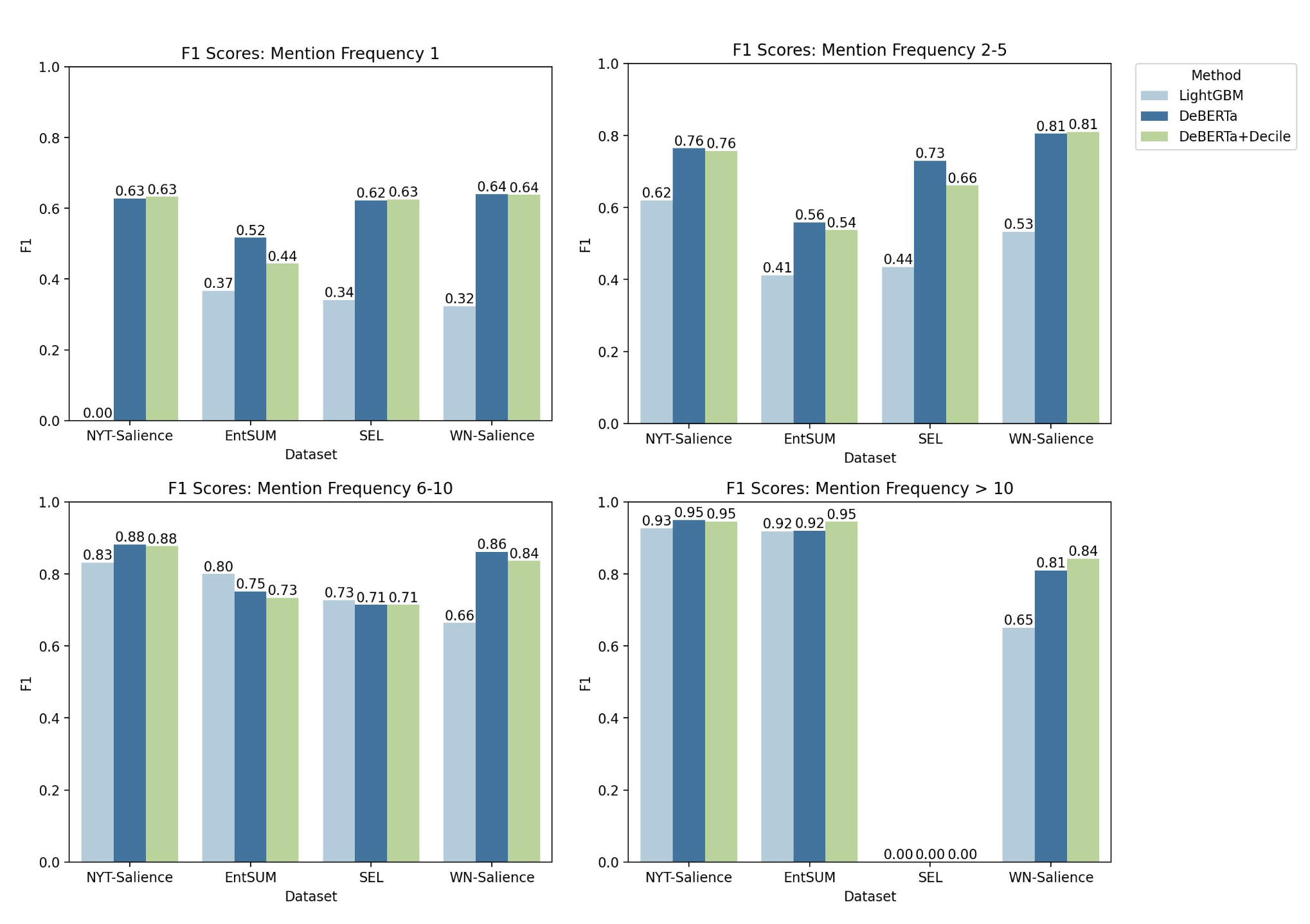}}\\
\caption{Stratified analysis across models and datasets.}
\label{fig:Analysis}
\end{figure*}

\subsection{Impact of Inferred Mentions}
\label{sec:impact-of-inferred-mentions}
In Section~\ref{sec:inferred mentions}, we inferred additional mentions of an entity for the NYT-Salience and WN-Salience datasets. We compare the performance of our best model that leverages multiple mentions of an entity to its version trained with only the first mentions of entities in a document. The specific input formats for this experiment are presented in Appendix~\ref{sec:input-format} The results in Table~\ref{tab:first_vs_all} show that doing so consistently improves the performance of our models across all datasets. In particular, for the largest dataset, NYT-Salience, our model achieves a substantial gain of 27.3 F1 points. This experiment showcases the importance of augmenting our datasets with additional mentions and the importance of explicitly modelling contextual information present around all entity mentions.

\subsection{Stratified Analysis on First Mention Position}
We compare our cross-encoder models against the Features \& GBDT model, our re-implemented baseline that relies on the most popular features used in prior works \cite{dunietz-gillick-2014-new,wu2020wn,trani-2018-sel}. As shown in the results from Tables~\ref{tab:results-nyt-wn} and~\ref{tab:results-sel-entsum}, among other features, positional features are most informative for salience. Intuitively, if an entity is mentioned in the headline or in the first sentence of a news article, there is high probability of that entity being salient. 

Figure~\ref{fig:Analysis} shows that all models perform well when the first mention falls in the headline or the first sentence of the document. We notice that the cross-encoder models constantly outperform the Features \& GBDT model and the largest gains are observed in the SEL and WN-Salience datasets. This observation indicates that the cross-encoder models are able to use the context to identify that mentions that occur in the headline or the first parts of the document are often salient without explicitly using this information as a feature.

We also investigate the performance of the models when the first mention falls inside or outside the context window of the PLM (here, 512 tokens). When mentions fall inside the context window, we observe that the cross-encoder models consistently outperform the Features \& GBDT model. When the mention falls outside the context window, the model predictions become close to random, which is expected, as the model does not have immediate contextual information around the mention.  Using models that can deal with longer inputs would be a promising direction for improvement for these samples~\cite{beltagy2020longformer}. Interestingly, for WN-Salience, the Features \& GBDT model also performs considerably worse outside the first 512 tokens. 

\subsection{Stratified Analysis on Mention Frequency}
Similar to mention position analysis, we compare our cross-encoder models against the Features \& GBDT model, which uses mention frequency as one of its input features. Figure ~\ref{fig:Analysis} shows how the cross-encoder models and Features \& GBDT compare with varying frequency of entity mentions.

For salient entities with single mentions, the cross-encoder model performs significantly better than the Features \& GBDT model. In particular, for the NYT-Salience dataset, the Features \& GBDT model fails to predict any of the single mention entities as salient. This observation indicates that the cross-encoder models do not simply model the mention frequency, but potentially leverage other contextual information to determine the salience of entities with a single mention. 

The performance of the Features \& GBDT model improves with more mentions per entity. In fact, for the frequency range of 6-10 mentions per entity, the Features \& GBDT model performs better than the cross-encoder models for EntSUM and SEL datasets. This observation indicates the over-reliance of the Features \& GBDT model on mention frequency to determine salience, but also that the cross-encoder cannot fully use this heuristic.

\section{Conclusion}

This paper aims to leverage the semantic knowledge encoded in pre-trained language models for entity salience detection. We propose the cross-encoder method based on Transformer-based PLMs with positional representations and compare its  performance to several ML-based methods, heuristic methods, and instruction-tuned LLMs across four different datasets, two human-annotated and two automatically curated. Across all our experiments, the cross-encoder model based on pre-trained language models outperforms all other methods, often with double digit gains in F-1 score. Analyses of model behavior illustrate the important effects of mention frequency, mention position, and document length on performance, highlighting areas of future work.

\section{Limitations}

We only studied salience in English-language documents, but our methods are applicable to other languages directly as long as a pre-trained language model covering the target language is available. 

We use entity mentions as annotated in our data or inferred through entity recognition and entity resolution for inference in some of the methods. This information may not be available at inference time in all applications. 

The experiments with LLMs are limited to zero-shot prompts. We did not experiment with instruction tuning which could potentially help the model learn the salience detection task.

Finally, we do not use external knowledge about entities and their relationships in modelling, which was shown to marginally improve results in past studies \cite{dunietz-gillick-2014-new,trani-2018-sel,ponza-2018-swat}. We consider this out of the scope of our analysis and a viable direction of future work.

\section{Ethics Statement}
We use publicly available datasets intended for the task of entity salience detection. The datasets and pre-trained models we used have permissive licenses allowing for research use. 
We do not envision any potential risks associated with the task discussed in this paper.

\bibliography{anthology,custom}

\appendix
\section*{Appendix}
\renewcommand{\thesubsection}{\Alph{subsection}}

\subsection{Details of Dataset Splits}
\label{sec:dataset split details}
\begin{table*}[t!]
    \centering
    \begin{tabular}{l|r|r|r|r}
    \toprule
    \textbf{Dataset} & \textbf{\# Doc-Entity pairs} & \textbf{Train} & \textbf{Validation} & \textbf{Test} \\
    \hline
    NYT-Salience & 1,910,214 & 1,342,092 & 405,335 & 162,787 \\
    WN-Salience & 62,537 & 41,625 & 11,902 & 9,009 \\
    SEL & 12,257 & 6,106 & 2,400 & 3,751 \\
    EntSUM & 9,934 & 5,206 & 1,861 & 2,867 \\
    \bottomrule
    \end{tabular}
    \caption{Document-Entity pairs in train, validation, and test splits after applying temporal splitting.}
    \label{tab:dataset-splits}
\end{table*}

Table~\ref{tab:dataset-splits} contains the train, dev, and test splits of each of the datasets after applying a temporal splitting strategy described in Section~\ref{sec:dataset splits}. These splits are used for model training and evaluation.

\subsection{Input Format for Experiments}
\label{sec:input-format}
As described in Section~\ref{sec:cross-encoder}, we add special marker tokens around each mention of the target entity (i.e., the entity for which the model needs to predict the salience label.). In the following, we provide an example:

\begin{tcolorbox}[colback=blue!5!white,colframe=blue!75!black,title=Text]
   Musk completes \$44 billion Twitter deal. Elon Musk, the world’s …
\end{tcolorbox}

\begin{tcolorbox}[colback=blue!5!white,colframe=blue!75!black,title=Model Input]
   [CLS] Elon Musk [SEP] [BEGIN\_ENTITY] Musk [END\_ENTITY] completes \$44 billion Twitter deal. [BEGIN\_ENTITY] Elon Musk [END\_ENTITY], the world’s …
\end{tcolorbox}

For the experiment with first mention reported in Section~\ref{sec:impact-of-inferred-mentions}, only the first mention is bounded by special marker tokens as shown in the following example:

\begin{tcolorbox}[colback=blue!5!white,colframe=blue!75!black,title=Model Input]
   [CLS] Elon Musk [SEP] [BEGIN\_ENTITY] Musk [END\_ENTITY] completes \$44 billion Twitter deal. Elon Musk, the world’s …
\end{tcolorbox}
Note that the second mention of \texttt{Elon Musk} is not bounded by marker tokens.

\subsection{Implementation details of zero-shot prompting of LLMs} \label{sec:zeroshot}

Figure~\ref{fig:instruction-llama-2} and Figure~\ref{fig:instruction-flan-ul2} show the prompts we used for the \texttt{LLaMa 2-Chat (7B)} and \texttt{Flan-UL2 (20B)} models respectively. Table \ref{gen-params} lists the generation parameters. We speculate the following causes for the relatively lower precision obtained using this method: \begin{itemize}
    \item The instruction defines the salience task definition, but doesn't provide any reference examples (few-shot prompting) to align with the definition of salience. This leads to the model identifying an entity as salient based on its frequency in the document. However, creating a few-shot prompt is challenging as we need to limit the maximum input length of the prompt to prevent out-of-memory issues.
    \item We truncate the document text so that the entire prompt is 2048 tokens or less, thus throwing away any potential information present towards the end of a long document.
\end{itemize}

\begin{figure}[ht!]
    \begin{mdframed}
        \begin{minipage}{\columnwidth} 

    <s> [INST] 
    \newline
    <<SYS>> 
    The salience of an entity provides information about the importance or centrality of that entity to the entire document text. In the following, given an Entity and a Text, you need to answer 'Yes' if the Text document is about that Entity and 'No'  if the Text is not about that Entity. <</SYS>>
    \newline
    Is Entity: \texttt{entity} salient in Text: \texttt{text}
    [/INST]
    \end{minipage}
    \end{mdframed}
    \caption{Instruction for zero-shot prompting of LLaMa 2-Chat model.}
    \label{fig:instruction-llama-2}
\end{figure}

\begin{figure}[ht!]    
    \begin{mdframed}
        \begin{minipage}{\columnwidth}                        
            \#\#\# Instruction \#\#\#
            \newline
            The salience of an entity provides infor-
mation about the importance or centrality of
that entity to the entire document text. In the
following, given an Entity and a Text, you need
to answer ’Yes’ if the Text document is about
that Entity and ’No’ if the Text is not about
that Entity.
            \newline
            Text: \texttt{text}
            \newline
            Entity: \texttt{entity}
            \newline
            Question: Is the above Entity salient in the above Text? Please answer Yes or No. 
            \newline
            Answer:             
        \end{minipage}
    \end{mdframed}
    \caption{Instruction for zero-shot prompting of Flan-UL2 model.}
    \label{fig:instruction-flan-ul2}
\end{figure}

\begin{table*}[ht!]
    \centering
    \begin{tabular}{|l|l|}
        \hline
        Generation parameter & Value                \\ \hline
        top\_k   & 0              \\ \hline
        top\_p & 0     \\ \hline
        temperature  & 0        \\ \hline
        max\_new\_tokens & 1 \\ \hline
    \end{tabular}
    \caption{Parameters for generating a salience label with zero-shot prompt.}
    \label{gen-params}
\end{table*}

\begin{figure*}[t!]
\includegraphics[scale=0.27]{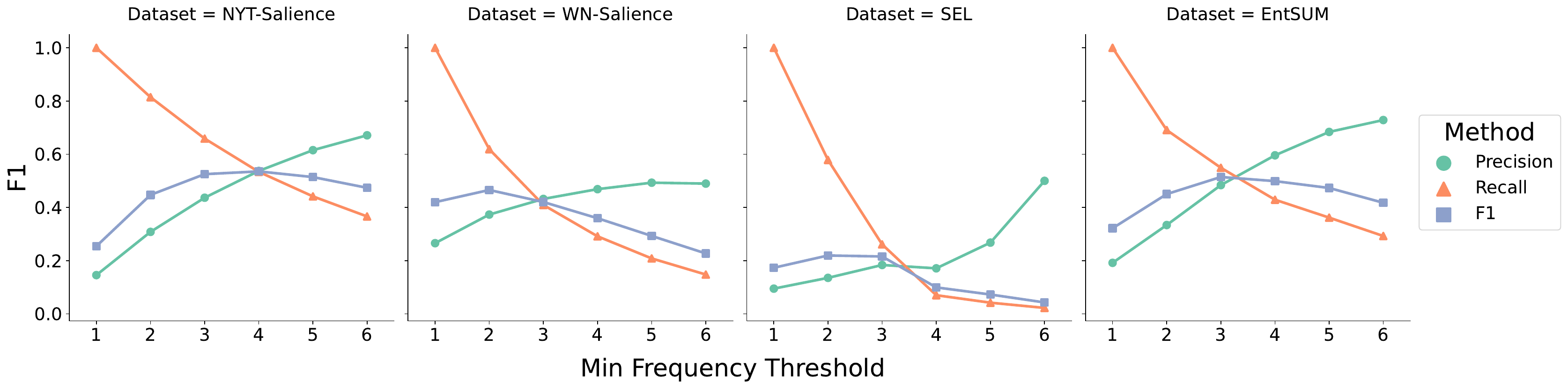}\caption{Performance of the Entity Frequency baseline over different thresholds.}\label{fig:frequency-threshold-plot}
\end{figure*}

\subsection{Thresholds for Entity Frequency baseline}

Figure~\ref{fig:frequency-threshold-plot} shows the performance of the Entity Frequency baseline by varying the minimum number of times an entity has to occur in the input document to be classified as salient.

\begin{figure*}[ht!]
    \centering
\includegraphics[scale=0.5]{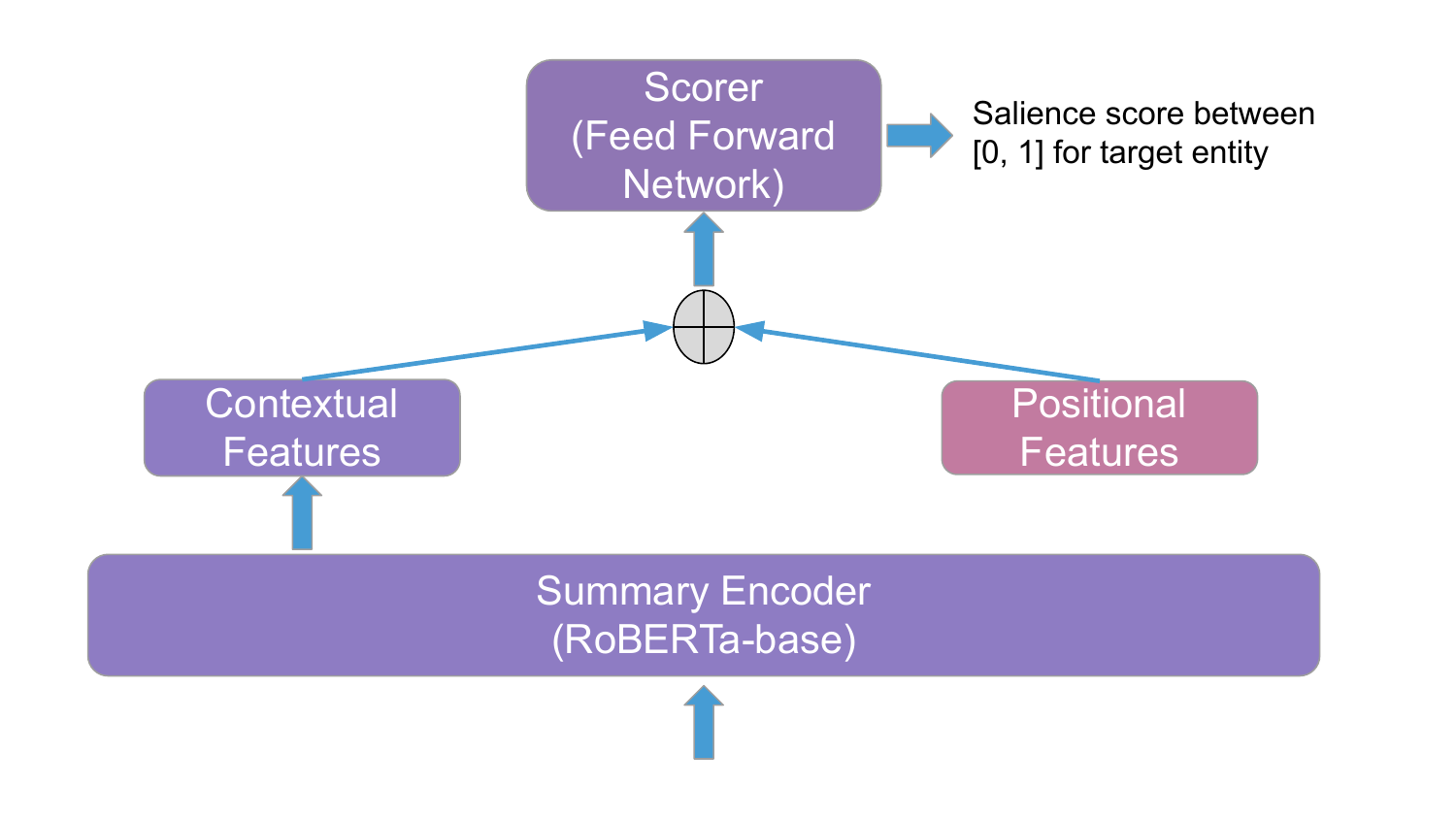}
\caption{Schematic diagram of the Target Entity Masking model architecture.}
\label{fig:target-masking}
\end{figure*}

\end{document}